# ChatGPT vs. Google: A Comparative Study of Search Performance and User Experience

Ruiyun (Rayna) Xu, Yue (Katherine) Feng, and Hailiang Chen[*]

July 2023

**Abstract**

The advent of ChatGPT, a large language model-powered chatbot, has prompted questions about its potential implications for traditional search engines. In this study, we investigate the differences in user behavior when employing search engines and chatbot tools for information-seeking tasks. We carry out a randomized online experiment, dividing participants into two groups: one using a ChatGPT-like tool and the other using a Google Search-like tool. Our findings reveal that the ChatGPT group consistently spends less time on all tasks, with no significant difference in overall task performance between the groups. Notably, ChatGPT levels user search performance across different education levels and excels in answering straightforward questions and providing general solutions but falls short in fact-checking tasks. Users perceive ChatGPT's responses as having higher information quality compared to Google Search, despite displaying a similar level of trust in both tools. Furthermore, participants using ChatGPT report significantly better user experiences in terms of usefulness, enjoyment, and satisfaction, while perceived ease of use remains comparable between the two tools. However, ChatGPT may also lead to overreliance and generate or replicate misinformation, yielding inconsistent results. Our study offers valuable insights for search engine management and highlights opportunities for integrating chatbot technologies into search engine designs.

**Keywords**: ChatGPT, generative AI, Google, search engines, chatbot, online experiment

[*] Xu is affiliated with Department of Information Systems and Analytics, Farmer School of Business, Miami University, Oxford, Ohio, USA. Feng is affiliated with Department of Management and Marketing, Faculty of Business, The Hong Kong Polytechnic University, Hong Kong. Chen is affiliated with Artificial Intelligence Research Institute, Faculty of Business and Economics, The University of Hong Kong, Hong Kong. Email: xur29@miamioh.edu, katherine.feng@polyu.edu.hk, and chen19@hku.hk. All authors contributed equally.

# 1. Introduction

In November 2022, OpenAI launched ChatGPT, a chatbot based on the Generative Pre-trained Transformer (GPT) large language model. ChatGPT's rapid rise in popularity underscores the transformative potential of generative AI in various industries and applications. In February 2023, Microsoft integrated ChatGPT into its Bing search engine, combining chat and search functionalities in a unique manner (Microsoft 2023). Following this integration, Bing experienced a significant traffic increase of 15.8% from February to March, while Google's traffic declined by nearly 1% during the same period (Reuters 2023). Considering that each 1% of search advertising market share represents $2 billion in annual revenue (Yahoo! Finance 2023), this notable shift raises concerns about the impact of ChatGPT-like products on traditional search engines and the future of search and information discovery.

Traditional search engines and ChatGPT-like systems differ in their information retrieval approaches. Google, the world's leading search engine, relies on keyword search and matching, presenting users with a list of relevant links. In contrast, ChatGPT employs a conversation-based approach, enabling users to pose queries in natural language. While Google's speed is impressive, users must filter through search results individually, which can be time-consuming. ChatGPT, however, aims to understand user intent and provide organized responses in complete sentences, offering a more user-friendly and intuitive search experience. Nevertheless, ChatGPT has potential drawbacks, such as slower response times and the possibility of false or misleading information, in contrast to traditional search engines that offer faster response times and more controlled results.

As the landscape of search and information discovery evolves, questions remain unanswered regarding the performance and user experience of ChatGPT-like systems compared to traditional search engines. It is crucial to examine how ChatGPT's conversational nature affects search result accuracy and relevance, explore the trade-offs between its intuitive responses and traditional search engines' faster results, and investigate how the transition from list-based results to conversational responses influences user satisfaction and search efficiency. Addressing these questions will aid in evaluating the advantages and disadvantages of adopting chat-based search systems like ChatGPT and guide the development of more effective search tools by revealing how user behaviors may evolve with the integration of AI-powered conversational systems.

In this study, we conduct a randomized online experiment with the objective of providing a comprehensive comparison between users' search behaviors and performance when utilizing large language model-powered chatbots and search engine tools, exemplified by ChatGPT and Google Search. We aim to answer the following research questions:

1) How do user behaviors differ when utilizing ChatGPT compared to Google Search for information-seeking tasks? Are these variations consistent across different types of search tasks?
2) Does ChatGPT level user search performance across different education levels?
3) How do users perceive the information quality, trust, and user experience of ChatGPT compared to traditional search engines like Google Search?

Two search tools are developed for the experiment to replicate the functions and interfaces of ChatGPT and Google Search, respectively. We recruit research participants through the Prolific platform (https://www.prolific.co/) and randomly assign them to one of our search tools to complete three information search tasks. After removing invalid responses, our final sample includes a total of 95 participants, with 48 in the ChatGPT group and 47 in the Google Search group. The results show that participants using ChatGPT consistently spend less time on all the tasks. However, their overall task performance does not differ significantly between the two tools.

Interestingly, we find that ChatGPT exhibits higher performance for tasks with straightforward questions but does not perform as well in fact-checking tasks, where we observe that ChatGPT is often incapable of correcting mistakes in the user prompt. We also find that ChatGPT levels user search performance across different education levels, while search performance on Google Search is positively correlated with education level. Participants perceive ChatGPT's responses to have higher information quality when compared to the information obtained from Google Search, despite displaying a comparable level of trust in both tools. Moreover, users in the ChatGPT group report significantly higher user experiences in terms of usefulness, enjoyment, and satisfaction, but a similar level of perceived ease of use.

As the first empirical study to systematically compare ChatGPT with traditional search engines like Google Search, this research makes several significant contributions to the academic literature on information search and human-computer interaction. By examining the differences in user behavior when utilizing search engines compared to chatbot tools, this study sheds light on how users adapt their information-seeking strategies to fit the affordances of these distinct technologies. Furthermore, our investigation into whether ChatGPT levels user search performance across different education levels enriches the literature on the digital divide by demonstrating the democratizing effects of advanced AI-based chatbot systems. Finally, by assessing users' perceptions of information quality, trust, and user experience in ChatGPT compared to traditional search engines, the research contributes to our understanding of user attitudes and preferences in the rapidly evolving landscape of information retrieval technologies. Importantly, the insights from this research inform the future development of large language models (LLMs) and search technologies, offering valuable guidance for creating more effective and user-centric tools in this field.

## 2. Literature Review

Our study aims to conduct a randomized online experiment to scrutinize the differences of search performance and experience exhibited by users when employing ChatGPT and the Google search engine for information retrieval. To this end, we undertake a thorough review of recent studies concerning ChatGPT and previous research on information search as well as experimental designs involving search users.

**2.1 ChatGPT and Its Impacts**

Recent advancements in LLMs, such as ChatGPT, have generated substantial interest due to their potential impact across various domains, including but not limited to research, education, finance, and healthcare. Many experts anticipate that LLMs will revolutionize these fields and lead to a paradigm shift. At the same time, concerns have emerged regarding potential issues associated with LLMs, such as hallucination, misinformation, copyright violation, institutionalizing bias, interpretability, misapplication, and over-reliance (Jo 2023; Sohail et al. 2023; Susarla et al., 2023).

Regarding the future of work, Eloundou et al. (2023) investigate the potential impact of LLMs on the U.S. labor market, suggesting that LLMs such as Generative Pre-trained Transforms exhibit traits of General-Purpose Technologies, leading to significant economic, social, and policy implications. Felten et al. (2023) present a methodology to estimate the impact of AI language modeling on various occupations and industries. Specifically, research conducted by MIT scholars demonstrates that ChatGPT significantly enhances productivity in professional writing tasks (Noy and Zhang, 2023).

In the realm of academic research, ChatGPT has demonstrated the capacity to transform research practices. Studies published in prestigious journals like Nature report that ChatGPT aids researchers in tasks such as analyzing and writing scientific papers, generating code, and facilitating idea generation (Dowling and Lucey 2023; Hustson 2023; Susarla et al., 2023; Van Dis et al. 2023).

In finance, ChatGPT has shown promise in forecasting stock price movements and improving the performance of quantitative trading strategies (Lopez-Lira and Tang, 2023). Hansen and Kazinnik (2023) investigate the ability of GPT models to decipher Fedspeak, specifically classifying the policy stance of Federal Open Market Committee announcements as dovish or hawkish. Wu et al. (2023) introduce BloombergGTP, a large language model with 50 billion parameters trained on both general-purpose and finance-specific datasets.

In the context of information search, there is a scarcity of research investigating how ChatGPT influences individuals' information-seeking behaviors compared to traditional search engines. To our knowledge, two medical studies have compared responses to health-related queries generated by ChatGPT and Google Search, finding that ChatGPT-generated responses are as valuable as, or even more valuable than, the information provided by Google (Hopkins et al., 2023; Van Bulck and Moons, 2023). However, these studies are limited in scope and represent medical experts' opinions. Our study differs from these studies in several ways. First, we focus on tasks in the general domain rather than the medical domain. Second, we conduct a randomized online experiment with many participants who perform searches by themselves and formulate their own queries. We also collect these search users' opinions and attitudes toward both tools. Lastly, we include an objective assessment of user search performance on both ChatGPT and Google Search.

**2.2 Information Search: Past and Present**

Internet search technologies have been continuously developing for over 30 years, starting with the creation of the first pre-Web Internet search engines in the early 1990s (Gasser 2006). In this section, we aim to present a concise review of search technologies, offering an overview of their evolutionary progress.

The first search engine, named Archie, was created in 1990 with the purpose of downloading directory listings from FTP sites and generating a database of filenames that could be easily searched (Gasser 2006). Subsequently, with the introduction of the World Wide Web in 1991, a wave of new search engines, such as Gopher, Veronica, and Jughead, emerged to assist users in navigating the rapidly expanding network. These early search engines were developed mainly based on indexing and keyword matching methods (Croft et al. 2010). In 1998, Larry Page and Sergey Brin, two Ph.D. students at Stanford University, developed a search engine algorithm called PageRank, which ranked web pages based on the number and quality of other pages linking to them (Brin and Page 1998). The PageRank approach revolutionized search by providing more relevant and high-quality results. This innovation laid the foundation for the establishment of Google, which rapidly emerged as the dominant search engine on a global scale, handling billions of search queries each day. Alongside Google, other popular search engines include Yahoo!, Bing, Baidu, Yandex and so on. The dominant search paradigm for most search engines is keyword-based search. Under this paradigm, a short query (e.g., a list of keywords) is submitted to a search engine, and the system then delivers relevant results. Relevant documents are selected mainly based on text matches, links, or domain information (Brin and Page 1998; Kleinberg 1999; Pokorny 2004).

Google has continuously strived to enhance search result quality and improve user experience through a series of launches and updates. In addition to numerous minor adjustments, the Google search engine has introduced more than 20 significant algorithmic updates (Search Engine Journal 2023). One notable update is the Panda algorithm, implemented in 2011, which introduced the concept of content quality as a ranking factor. The Panda algorithm evaluates factors such as originality, authority, and trustworthiness to assess the quality of web pages (Goodwin 2021). Machine learning algorithms play a crucial role in assigning pages a quality classification that aligns with human judgments of content quality. In 2012, Google launched the Penguin algorithm, further bolstering search quality. This update specifically targeted web spam by identifying and devaluing pages that employed black hat link building techniques to artificially boost their rankings (Schwartz 2016). By penalizing such manipulative practices, the Penguin algorithm aimed to ensure that search results prioritized high-quality and relevant content.

The state-of-the-art search technologies utilize artificial intelligence and knowledge graphs. For instance, back in 2012, Google announced a Google Knowledge Graph covering a wide swath of subject matter and applied the knowledge graph to enable more intelligent search by providing instant answer to users' search queries (Google 2012). Google Hummingbird, an algorithm launched in 2013, is an AI-based system that helps understand the context and meaning behind search queries. The Hummingbird algorithm has advanced beyond the basic practice of matching keywords in a query to keywords on a webpage. Instead, it has become more sophisticated in its ability to present pages that closely align with the inherent topic of the search query (Montti 2022). Additionally, the Hummingbird algorithm enables processing of longer conversational search queries.

The potential of generative AI to revolutionize information search is immense due to its great performance in natural language understanding. However, there is still much to explore regarding how this cutting-edge technology impacts users' search performance and experience. Understanding these effects is crucial for harnessing the full potential of generative AI to enhance user experience in information search. In this study, we delve into this domain and strive to provide a comprehensive comparison between traditional search engines and ChatGPT, shedding light on their respective strengths and capabilities.

## 2.3 Experimental Design for Search Users

To examine how users interact with search engines to address various search tasks and how an enhanced search engine design improves user performance, researchers often employ experimental methods that involve simulating realistic search scenarios. These methods allow researchers to observe and analyze users' search behavior and performance in controlled experimental settings (e.g., Adipat et al. 2011; Liu et al., 2020; Sagar et al., 2019; Storey et al. 2008). In our study, we build upon these prior studies by designing a range of search tasks for our experiment. We manipulate the task complexity, drawing inspiration from studies such as Wildemuth and Freund (2004) and Liu et al. (2020). Additionally, we incorporate commonly used metrics to assess search performance and user experience, including time spent on search tasks, task performance, perceived information quality, satisfaction, and others (Sargar et al. 2019, Liu 2021).

## 3. Experimental Design and Data

We employ a between-subjects design with two conditions (LLM-powered chatbot vs. traditional search engine) in the online experiment, where participants are randomly assigned to one of the two conditions. For this purpose, we develop two website tools: one simulating ChatGPT and the other mimicking Google Search. To ensure a realistic user experience, we closely replicate the interfaces of ChatGPT and Google Search. Figures 1 and 2 display screenshots of the interfaces for each tool, respectively. For the chatbot tool, we employ OpenAI's Chat Completion API (https://platform.openai.com/docs/api-reference/chat) and the gpt-3.5-turbo model to generate responses to user prompts. Both user prompts and API responses are displayed on the same webpage for each chat session. The chatbot retains memory of the past few rounds of prompts and responses, allowing it to conduct natural conversations. For the search engine tool, we utilize Google's Custom Search JSON API (https://developers.google.com/custom-search/v1/overview) to handle search queries. The search results are displayed on different pages, with each page containing at most 10 result items. To monitor user search behavior, we provide each participant with a pre-registered user account. Participants must log in to their assigned accounts and use the corresponding tool to perform their tasks. For the ChatGPT tool, we record each user prompt and the corresponding response generated by the GPT-3.5 model. For the Google Search tool, we track each submitted search query, the different page views of search results for the same query, and any clicks on search result items. The timestamps of these user and API actions are also recorded.

We recruit 112 participants through the Prolific platform, using the selection criteria of being located in the USA and having English as their first language. Participants are also required to use desktop computers to complete the study. These participants are then randomly assigned to either the ChatGPT or Google Search group for the experiment. Participants are instructed to use the assigned tool to complete three tasks, and the use of any other tools is strictly prohibited. Furthermore, we require participants to avoid relying on their own knowledge to provide answers. We also ask participants to record the time they spend on each task using a timer. To ensure that participants clearly understand the requirements and instructions, we include two comprehension questions before they can proceed with the tasks. In addition, participants who fail attention checks during the process are removed from our sample. Our final sample comprises 95 participants, with 48 using the ChatGPT tool and 47 using the Google Search tool for the tasks.

We design three tasks with varying levels of complexity for the experiment by referring to previous research on search user experiment. Specifically, Task 1 involves a specific question that asks participants to find out "the name of the first woman to travel in space and her age at the time of her flight" (Wildemuth and Freund 2004). In Task 2, participants are required to list five websites with links that can be used for booking a flight between two cities (Phoenix and Cincinnati) in the USA. Task 3 is a fact-checking task in which we ask participants to read an excerpt of a news article and fact-check three italicized statements: (1) *The 2009 United Nations Climate Change Conference, commonly known as the Copenhagen Summit, was held in Copenhagen, Denmark, between 7 and 15 December*; (2) *On the final day of the conference, the UN climate summit reached a weak outline of a global agreement in Copenhagen, which fell significantly below the expectations of Britain and many poor countries*; and (3) *The United States drew substantial criticism from numerous observers as they arrived at the talks with a proposal of a mere 6% reduction in emissions based on 1990 levels*. Participants need to indicate whether each statement is "True" or "False" and provide evidence or corrections, if any. The design of Task 3 draws inspiration from a prior study conducted by Liu et al. (2020), but the specific details have been developed from scratch.

Each participant is given a link to the randomly assigned tool, as well as a username and password to use the tool. After searching and obtaining results from the tool, they need to provide their answers on our study webpage hosted on Qualtrics. Later, we check the accuracy of these submitted answers to assess the search performance of participants. There are standard answers to Tasks 1 and 3. Although there can be many different correct answers to Task 2, each answer can be easily checked by visiting the provided links and verifying their contents. To accomplish this, we employ two research assistants (RAs) who manually and independently verify whether each link submitted in Task 2 points to a flight booking website, is a valid web link, directs to the homepage only, or displays a flight between Phoenix and Cincinnati. In cases where the two RAs disagree, one of the co-authors steps in as a tiebreaker and makes the final judgment call.

After participants complete the tasks, we ask them to fill out a questionnaire and collect their perceptions of ease of use, usefulness, enjoyment, and satisfaction with using the tool. We also collect their perceived information quality of the tool's responses and trust in using the tool. Moreover, we check the manipulations by asking participants about the features of the assigned tool. At the end of the questionnaire, we collect the participants' background information (e.g., age, gender, level of education, etc.), their prior experience with ChatGPT and search engines, and their prior knowledge on the topics of the given search tasks. The detailed measurements are illustrated in the Appendix.

## 4. Results

In this section, we present the results of our experiment based on the analyses of participants' task responses, questionnaire answers, and search behaviors tracked through server logs. Table 1 reports the results of manipulation and randomization checks. Table 2 presents comparisons between the two experimental groups (ChatGPT vs. Google Search) regarding search efficiency, efforts, performance, and user experience. Analysis of Variance (ANOVA) is performed to evaluate whether the differences between the two groups are significant.

### 4.1 Manipulation and Randomization Checks

We first conduct a manipulation check to ensure that our manipulation is successfully implemented and that our experimental design is valid. As shown in Panel A of Table 1, participants in the ChatGPT group believe that they use a tool that is significantly different from a traditional search engine and features a conversational interface (5.61 vs. 4.64, $p<0.05$). The manipulation check questions utilize a 7-point scale, with higher scores indicating participants' belief that the search tool used in the task has a conversational interface and differs from traditional search engines.

Since some participants fail the attention checks and are consequently removed from our sample, we also verify the validity of the randomization for our final sample by comparing the participants' demographics (age, gender, level of education, employment status), prior knowledge of the topics, and prior experience with relevant technologies. The results in Panel B of Table 1 confirm that there are no significant differences between the two groups in these aspects.

**4.2 Search Efficiency**

We begin by comparing the two search tools, focusing on participants' search efficiency, quantified as time spent on each task (including providing answers) and using the search tool. Panel A of Table 2 reports the comparison results between the two experimental groups. Notably, we employ two approaches to measure the time: self-reported task time by participants and objective time spent using the tool retrieved from server logs.

Based on self-reported results, on average, it takes participants in the ChatGPT group 11.35 minutes to complete the three tasks, while it takes those in the Google Search group 18.75 minutes (i.e., 65.20% more). Across all three tasks, the ChatGPT group consistently spends much less time on each task than the Google Search group. All these differences are statistically significant at the 1% level.

Furthermore, we analyze the server logs of the two search tools to calculate the time spent on each task in an objective way. For the ChatGPT group, time spent on search is measured by the time span from the user's initial query to the last response received from the ChatGPT API. For the Google Search tool, we use the duration between the user's first query and their last click to capture the time spent. If the last query is not followed by any click, the end time is the user's last query. It is worth noting that while the server-log measures are more objective, they are likely an underestimate of the true time spend on the tool because the server log does not record the exact moment when participants finish a task or leave the tool. In addition, the server-log measures are also likely to exclude the time for participants to refine their answers and fill out the questionnaire. As such, the time spent calculated from server logs is relatively less than the time reported by the participants themselves. Nevertheless, we use this time retrieved from server logs as an alternative measure to cross-validate the search efficiency between the two tools, in addition to participants' self-reported time. The results in the lower part of Panel A in Table 2 suggest a consistent pattern such that the time spent on the ChatGPT is significantly less than that on Google Search across all three tasks.

We attribute the observed difference in search efficiency between ChatGPT and Google Search to the distinct ways in which users interact with and obtain information from these tools. When using Google Search, users must formulate search queries on their own, often going through a trial-and-error process to find the most relevant results. This can be time-consuming, as users need to sift through search results, sometimes relying on luck to find the desired information. On the other hand, ChatGPT allows users to simply ask a question in natural language, streamlining the search process. ChatGPT then provides a summarized answer to the user's question, eliminating the need for additional research or reading. This more direct method of obtaining information enables users to find answers more efficiently, resulting in significantly less time spent for the ChatGPT group compared to the Google Search group. Remarkably, our findings based on server logs reveal that the average time spent on Task 1 and Task 2 using the ChatGPT tool is less than one minute, suggesting that participants made only a limited number of queries and were able to obtain answers directly from ChatGPT. This further underscores the efficiency of ChatGPT in providing immediate responses to users, especially in search tasks with specific and clear information needs.

## 4.3 Search Efforts

We examine the user prompts and search queries extracted from server logs to understand how users interact with the AI-powered chatbot and search engine. Specifically, we focus on how the participants formulate queries during search tasks, as indicated by the average number of queries for each task and the average length of their queries. The results presented in Panel B of Table 2 show that participants in the ChatGPT group use a similar number of queries across three tasks as those in the Google Search group, but the average query length is significantly larger for the ChatGPT group.

For the number of queries, participants in the ChatGPT group use significantly fewer search queries (i.e., user prompts) to complete the first task compared to those in the Google Search group (1.55 vs. 2.13, $p<0.01$). In Task 2, while participants in the ChatGPT group still use a relatively smaller number of queries, the difference is minor and marginally significant at the 10% level. Task 2 involves compiling a list of websites with links, which is a task that Google is well-suited for. Therefore, participants in both groups can complete the task with minimal effort, using fewer than two queries on average. Conversely, for the more complex Task 3, there is no significant difference between the two search tools, although participants in the ChatGPT group conduct slightly more queries than those in the Google Search group.

Regarding query length, our findings suggest that ChatGPT users tend to formulate significantly longer queries in search tasks compared to Google Search users. The results show that the query length from participants in the ChatGPT group is consistently greater across all three tasks than those in the Google Search group. This is likely because ChatGPT is designed to engage in natural language conversations with users. In contrast to Google Search, which requires short and concise keyword input, ChatGPT allows users to interact in a more conversational manner.

Consequently, users may feel more comfortable using longer, natural language queries and providing additional context and details about their inquiries when interacting with ChatGPT. Our findings highlight the need for users to adapt their search habits due to the unique conversational search paradigms employed by ChatGPT, as opposed to the keyword-centric design of traditional search engines.

**4.4 Search Performance**

To assess search performance, we evaluate each participant's answer to each task using a scoring system based on a total of 10 points. Each task in our experiment has objective answers. For example, the correct answers for Task 1 are Valentina Tereshkova (the name of the first woman to travel in space) and 26 years old (her age at the time of her flight). Participants earn 5 points for each correct answer, achieving a full score of 10 points if both answers are correct. Similarly, in Task 2, participants can get 2 points for each correct answer out of five websites with links. In Task 3, participants are required to check three statements and provide evidence. We assign equal weight to each check, such that participants can earn 10/3 points for each correct answer. This scoring system allows us to compare the search performance between the ChatGPT and Google Search groups effectively.

Panel C of Table 2 presents the comparison results for search performance. On average, participants in the ChatGPT group score a total of 8.55, while participants in the Google Search group score 8.77. The difference between the two groups is only -0.22, which is statistically insignificant at the 10% level. These findings are particularly noteworthy, considering that the Google Search group spends 65.2% more time (as demonstrated in our earlier analysis) to achieve a same level of performance. The implications of this are substantial, as it suggests that ChatGPT can significantly enhance the productivity of search users while maintaining the same level of task performance.

While there is no significant difference in the overall task performance between the two experimental groups, a detailed comparison reveals varying performances across individual tasks. Notably, in Task 1, all participants using ChatGPT achieve full marks, displaying superior performance and suggesting that ChatGPT is highly effective at fact retrieval. In contrast, Google Search users make several errors, with an average score of 8.19. The difference of 1.81 is statistically significant at the 1% level. Although the first search result provided by Google contains the correct answer to Task 1, participants still need to read through the result page to find the correct information. Due to multiple names mentioned in the article, participants often mistakenly use the wrong name. Consequently, the ChatGPT group's performance is significantly better than the Google Search group's in Task 1.

We further examine the participants' task performance between the two groups across different education levels. Intriguingly, as illustrated in Figure 3, we observe that participants holding a doctorate degree show no significant difference in answering Task 1, regardless of whether they use ChatGPT or Google Search. However, the performance of Google Search users with other educational backgrounds is consistently lower than that of ChatGPT users. This result aligns with a recent finding by Noy and Zhang (2023) that ChatGPT helps reduce inequality between workers by benefiting low-ability workers more. In our study, participants exhibit the same performance in Task 1, irrespective of their educational backgrounds, when using ChatGPT, while the performance of Google Search users largely depends on their education levels. Based on Figure 3, we infer that using Google Search tends to be more challenging for users with lower levels of education.

We observe no significant difference in performance between the two experimental groups in Task 2. Given that Task 2 requires a list of websites and links, it is important to note that Google Search's default output inherently presents a list of relevant websites and links, while ChatGPT excels at providing summarized responses. As a result, both tools demonstrate comparably exceptional performance, as indicated by average scores that are close to the full mark (9.81 and 9.74, respectively). Upon examining the link details more closely, we find that most links provided by participants in the ChatGPT group are homepages of flight booking websites, while the websites provided by participants in the Google Search group specifically direct to flights between the two cities (i.e., from Phoenix to Cincinnati) as required in the task. If we consider answers specifying the correct flight departure and destination as the criterion for performance evaluation, the Google Search group performs significantly better than the ChatGPT group (8.88 vs. 5.00, $p<0.01$). Considering that users typically need to provide specific keywords in Google Search, it is more likely to yield targeted and specific results compared to the more general responses generated by ChatGPT. We further examine the performance distributions based on participants' educational backgrounds. As illustrated in Figure 4, no significant differences are evident between the two groups across various education levels. Notably, participants holding master's and doctorate degrees achieve perfect scores in both groups.

In contrast, user performance in Task 3 (the fact-checking task) is significantly worse in the ChatGPT group than in the Google Search group (5.83 vs. 8.37, $p<0.01$). Examining the responses from ChatGPT reveals that it often aligns with the input query, replicating inaccuracies in subsequent responses. For instance, when we enter the prompt "*Is the following statement true or false? 'The 2009 United Nations Climate Change Conference, commonly known as the Copenhagen Summit, was held in Copenhagen, Denmark, between 7 and 15 December.'*" to ChatGPT, it responds to us with "*The statement is true. The 2009 United Nations Climate Change Conference was indeed held in Copenhagen, Denmark, between 7 and 15 December.*" The accurate conference dates are between 7 and 18 December. Surprisingly, after we change our prompt to "*When was the 2009 UN climate change conference held?*", ChatGPT provides the correct answer. More importantly, participants often demonstrate a lack of diligence when using ChatGPT and are less motivated to further verify and rectify any misinformation in its responses. According to our observations, 70.8% of the participants in the ChatGPT group demonstrate an overreliance on ChatGPT responses by responding with "True" for the first statement. While evaluating the accuracy of the third statement in Task 3, we observe that ChatGPT tends to offer inconsistent answers for the same prompt during multiple trials. Furthermore, although it occasionally recognizes the statement as incorrect, it fails to provide accurate information (i.e., the exact percentage of emission reduction). Similar to Tasks 1 and 2, we examine the distributions of Task 3 performance across different education levels. As depicted in Figure 5, participants in the ChatGPT group consistently have lower performance than Google Search users across all education levels. The performance of the ChatGPT group does not vary with participants' education backgrounds. By contrast, performance with Google Search is positively related to users' education levels. Users with advanced education levels demonstrate greater proficiency in using Google Search to correct mistakes in the fact-checking task.

### 4.5 User Experience

The data collected from the questionnaire provides additional support for the aforementioned arguments. Our results in Panel D of Table 2 show that participants in the ChatGPT group perceive the information in the responses to be of considerably higher quality than those in the Google Search group (5.90 vs. 4.62, $p<0.01$). ChatGPT delivers organized responses in complete sentences to users' queries, potentially making the information more accessible. However, we do not identify a significant difference in participants' trust in using these two tools. Participants tend to accept the responses as provided and exhibit a lack of inclination to question the information sources from both tools. While participants display a similar level of trust in using both tools, Google Search users may need to exert more effort and spend additional time browsing webpages to locate relevant information. Therefore, their perceived information quality is lower. In contrast, ChatGPT's convenience may discourage participants from further exploring and verifying information in its responses, resulting in subpar performance in fact-checking tasks. In addition, participants in the ChatGPT group find it to be more useful and enjoyable and express greater satisfaction with the tool compared to those in the Google Search group. Perceived ease of use is relatively higher in the ChatGPT group than in the Google Search group, but the difference is not significant at the 5% level. This may be attributed to people's existing familiarity with Google, and the tasks in our experiments may not pose a significant challenge for them.

## 5. Discussion and Conclusion

This study provides a comprehensive comparison of search performance and user experience between ChatGPT and Google Search. By conducting a randomized online experiment, the research highlights the trade-offs between the conversational nature of ChatGPT and the list-based results of traditional search engines like Google. On one hand, the utilization of ChatGPT has shown considerable enhancements in work efficiency, enabling users to accomplish tasks in less time, and can foster a more favorable user experience. On the other hand, it is important to note that ChatGPT does not always outperform traditional search engines. While ChatGPT excels in generating responses to straightforward questions and offering general solutions, this convenience may inadvertently hinder users from engaging in further exploration and identifying misinformation within its responses. The findings based on our survey questions further support that people believe the information generated by ChatGPT has higher quality and is more accessible than Google Search, and they hold a similar view of trust in both results. Interestingly, our findings suggest that ChatGPT has a leveling effect on user performance, regardless of their educational backgrounds, while users with higher levels of education display more proficiency in using Google Search.

As users increasingly seek more efficient and user-friendly search tools, the integration of AI-powered conversational systems like ChatGPT can significantly impact the search engine market. Businesses and search engine providers must consider the advantages and disadvantages of adopting chat-based search systems to enhance search efficiency, performance, and user experience. Future research should explore other types of search tasks and gain a deeper understanding of how users interact with AI-powered conversational systems differently from traditional search engines. It is also important to investigate the long-term effects of adopting such systems on search behaviors and the search engine market. Lastly, future studies could examine the integration of chat and search functionalities and explore the optimal balance between conversational and keyword-based approaches.

Pokorný, J. 2004. "Web Searching and Information Retrieval," *Computing in Science & Engineering* (6:4), pp. 43-48.

Reuters. (2023). OpenAI tech gives Microsoft's Bing a boost in search battle with Google. https://www.reuters.com/technology/openai-tech-gives-microsofts-bing-boost-search-bat

Schwartz (2016). Google updates Penguin, says it now runs in real time within the core search algorithm. *Search Engine Land.* Accessed on June 25, 2023.

Seach Engine Journal (2023). History of Google Algorithm Updates. Accessed on June 25, 2023.

Sohail, S. S., Farhat, F., Himeur, Y., Nadeem, M., Madsen, D. Ø., Singh, Y., Atalla, S. & Mansoor, W. (2023). The future of GPT: A taxonomy of existing ChatGPT research, current challenges, and possible future directions. *Working Paper.*

Storey, V. C., Burton-Jones, A., Sugumaran, V., and Purao, S. 2008. "CONQUER: Methodology for Context-Aware Query Processing on the World Wide Web," *Information Systems Research* (19:1), pp. 3-25.

Susarla, A., Gopal, R., Thatcher, J. B., & Sarker, S. (2023). The Janus effect of generative AI: Charting the path for responsible conduct of scholarly activities in information systems. *Information Systems Research*, Articles in Advance, 1-10.

Yahoo! Finance. (2023). Microsoft's Bing is the first threat to Google's search dominance in decades. Retrieved from https://finance.yahoo.com/news/microsofts-bing-is-the-first-threat-to-googles-search-dominance-in-decades-210913597.html

Van Dis, E. A., Bollen, J., Zuidema, W., van Rooij, R., & Bockting, C. L. (2023). ChatGPT: Five priorities for research. *Nature*, 614(7947), 224-226.
23

Wu, S., Irsoy, O., Lu, S., Dabravolski, V., Dredze, M., Gehrmann, S., Kambadur, P., Rosenberg, D., & Mann, G. (2023). BloombergGPT: A Large Language Model for Finance. arXiv:2303.17564. Working Paper.

Van Bulck, L., & Moons, P. (2023). What if your patient switches from Dr. Google to Dr. ChatGPT? A vignette-based survey of the trustworthiness, value, and danger of ChatGPT-generated responses to health questions. *European Journal of Cardiovascular Nursing*, 00, 1-4.24

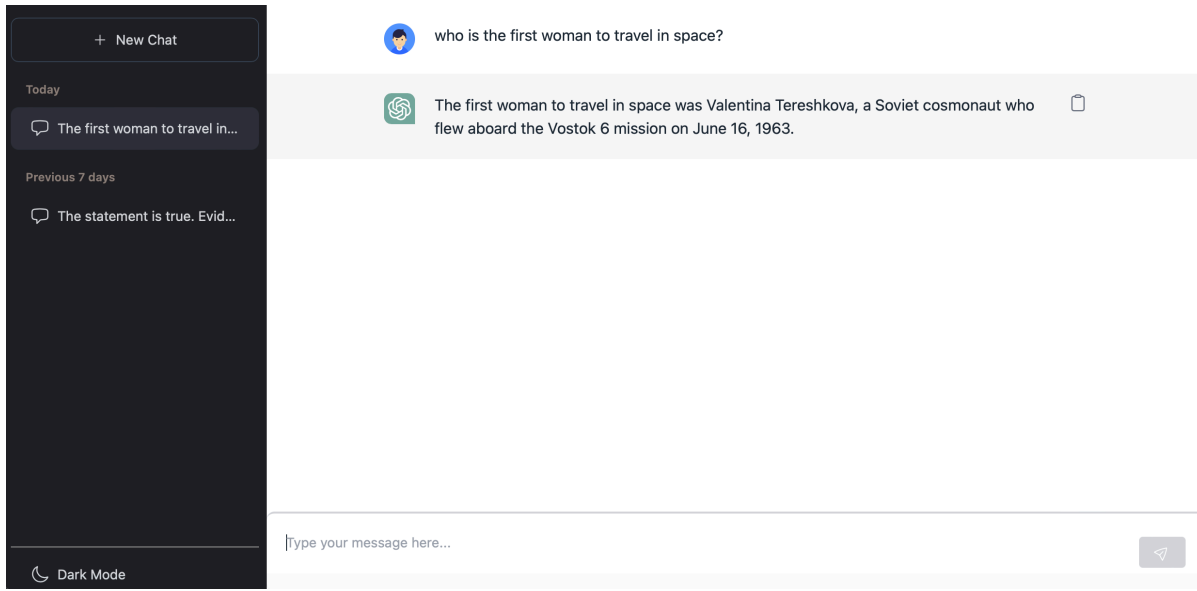

**Figure 1. Screenshot of the Chatbot Tool (ChatGPT)**

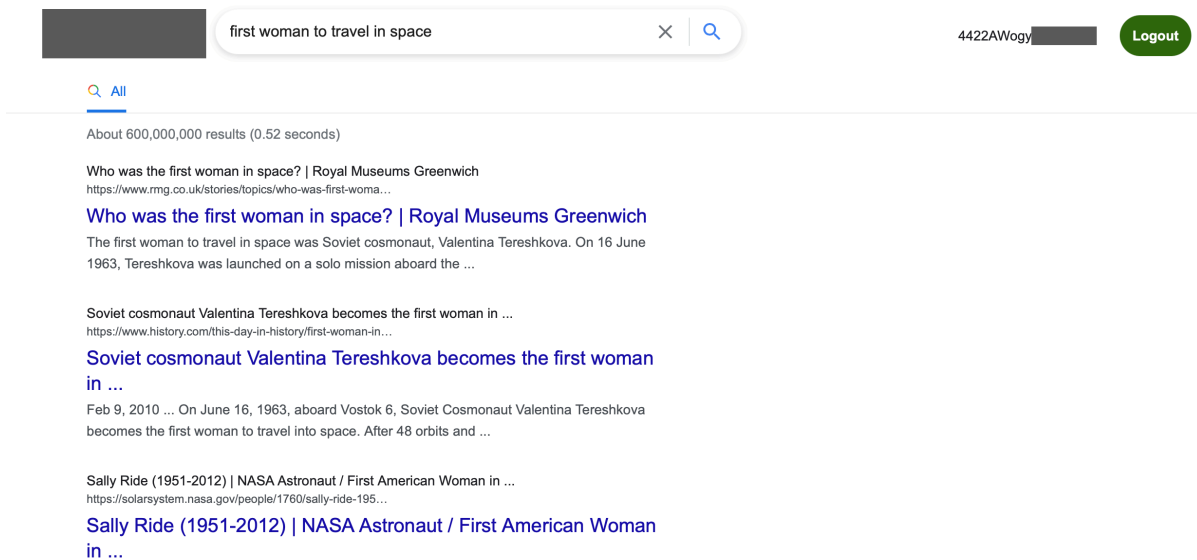

**Figure 2. Screenshot of the Search Tool (Google Search)**

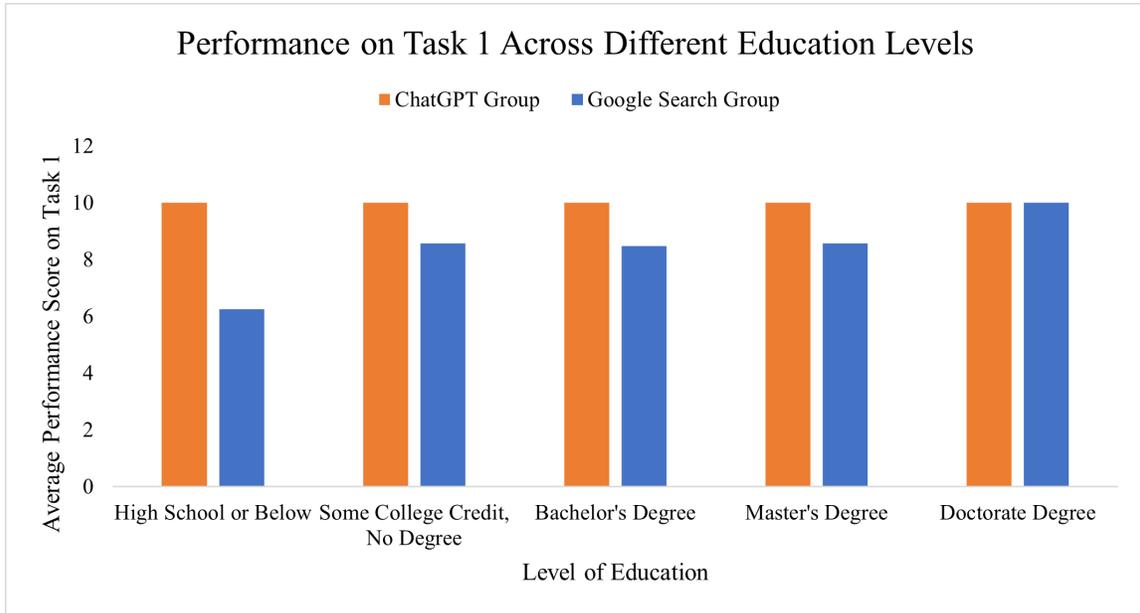

**Figure 3. Comparisons of Performance on Task 1 between Two Groups across Education Levels**

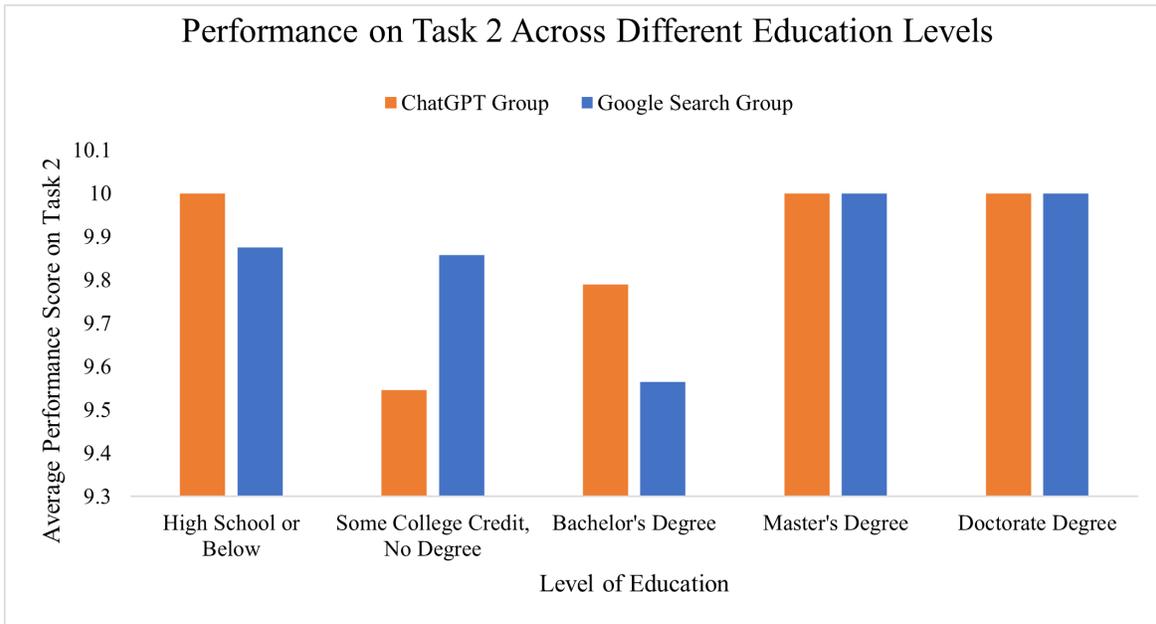

**Figure 4. Comparisons of Performance on Task 2 between Two Groups across Education Levels (General Criterion)**

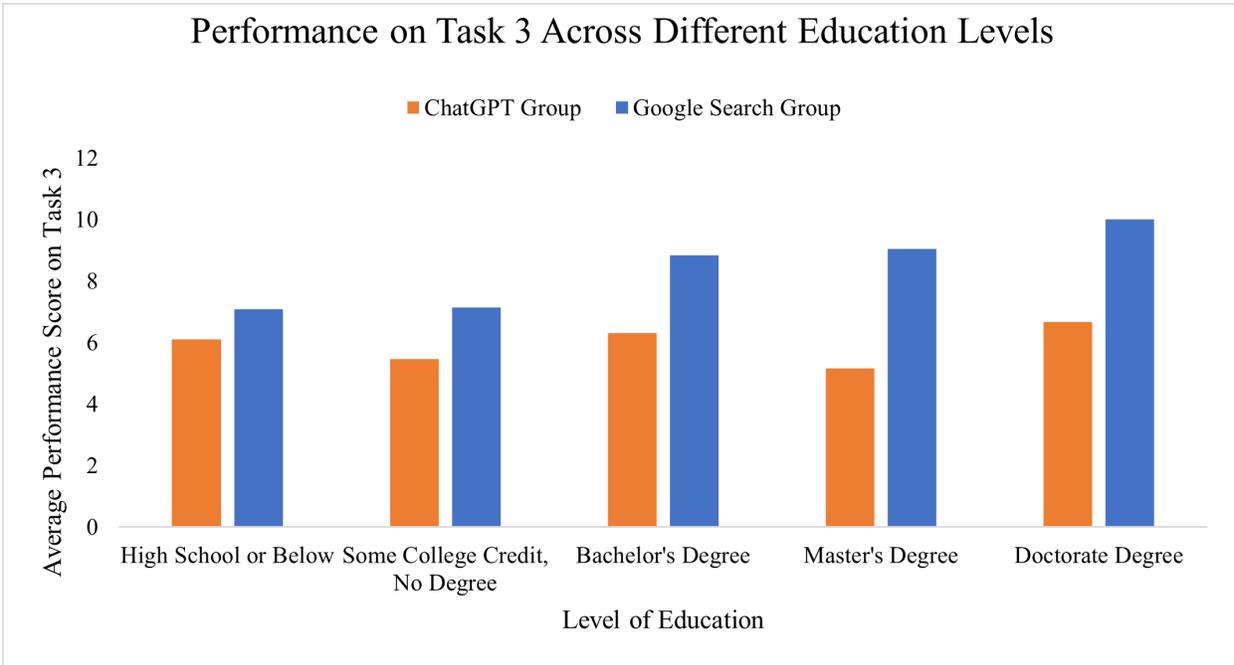

**Figure 5. Comparisons of Performance on Task 3 between Two Groups across Education Levels**

**Table 1. Manipulation and Randomization Checks**

| Measure | ChatGPT (48 participants) | Google Search (47 participants) | Difference (ChatGPT – Google) | F-statistic |
|---|---|---|---|---|
| **Panel A. Manipulation Check** | | | | |
| Perceived features of the assigned tool | 5.61 | 4.64 | 0.98 | 6.77** |
| **Panel B. Randomization Check** | | | | |
| Age | 3.00 | 3.23 | -0.23 | 1.88 |
| Gender | 1.40 | 1.30 | 0.10 | 0.72 |
| Education level | 2.79 | 2.74 | 0.05 | 0.05 |
| Employment Status | 2.13 | 1.85 | 0.27 | 0.84 |
| Familiarity with topics in the tasks | 3.56 | 4.09 | 0.52 | 1.97 |
| Prior experience with search engines | 4.98 | 4.98 | 0.00 | 0.00 |
| Frequency of search engine usage | 1.08 | 1.17 | -0.09 | 1.17 |
| Self-rated search skill | 3.00 | 2.98 | 0.02 | 1.02 |
| Prior experience with ChatGPT | 2.83 | 3.32 | -0.49 | 1.68 |

Notes: *(1) Analysis of variance (ANOVA) is employed to test the difference between the two groups. (2) Significant level: \*\*\* p < 0.01; \*\* p < 0.05; \* p < 0.1.*

**Table 2. Comparisons of Search Performance, Behavior, and User Experience**

| Measure Group Mean | ChatGPT (48 participants) | Google Search (47 participants) | Difference (ChatGPT - Google) | F-statistic |
|---|---|---|---|---|
| **Panel A. Search Efficiency** | | | | |
| **Self-reported Task Time (min)** | | | | |
|   Total time for three tasks | 11.35 | 18.75 | -7.40 | 26.88*** |
|   Time spent on task 1 | 1.83 | 3.37 | -1.54 | 18.09*** |
|   Time spent on task 2 | 2.40 | 3.61 | -1.20 | 7.22*** |
|   Time spent on task 3 | 7.12 | 11.78 | -4.66 | 22.86*** |
| **Time Spent on Search Tool (min)** | | | | |
|   Total time for three tasks | 5.79 | 14.95 | -9.15 | 34.81*** |
|   Time spent on task 1 | 0.34 | 2.42 | -2.08 | 22.11*** |
|   Time spent on task 2 | 0.52 | 2.78 | -2.26 | 40.39*** |
|   Time spent on task 3 | 4.93 | 9.81 | -4.88 | 14.06*** |
| **Panel B. Search Efforts** | | | | |
| Total # of queries on three tasks | 7.36 | 8.13 | 0.77 | 1.30 |
| # of queries (task 1) | 1.55 | 2.13 | -0.58 | 7.13*** |
| # of queries (task 2) | 1.30 | 1.65 | -0.35 | 3.39* |
| # of queries (task 3) | 4.51 | 4.35 | 0.16 | 0.09 |
| Average query length for three tasks | 37.54 | 12.05 | 25.49 | 27.59*** |
| Query length (task 1) | 13.50 | 9.90 | 3.60 | 12.84*** |
| Query length (task 2) | 18.43 | 6.11 | 12.32 | 156.63*** |
| Query length (task 3) | 80.72 | 19.82 | 60.90 | 18.74*** |
| **Panel C. Search Performance (Full Score: 10)** | | | | |
| Average performance score on three tasks | 8.55 | 8.77 | -0.22 | 0.83 |
| Performance score on task 1 | 10.00 | 8.19 | 1.81 | 19.46*** |
| Performance score on task 2 | 9.81 | 9.74 | 0.07 | 0.14 |
|   *If answers pointing to the destinations* | 5.00 | 8.88 | -3.88 | 201.68*** |
| Performance score on task 3 | 5.83 | 8.37 | -2.54 | 24.23*** |
| **Panel D. User Experience** | | | | |
| Perceived information quality | 5.90 | 4.62 | 1.27 | 15.85*** |
| Technology trust | 5.38 | 5.30 | 0.07 | 0.46 |
| Perceived ease of use | 6.00 | 5.57 | 0.43 | 3.80* |
| Perceived usefulness | 6.19 | 5.30 | 0.89 | 10.10*** |
| Perceived enjoyment | 5.87 | 4.74 | 1.12 | 14.19*** |
| Satisfaction | 6.06 | 5.27 | 0.79 | 9.32*** |

*Note: (1) Analysis of variance (ANOVA) is employed to test the difference between the two groups.*
*(2) Significant level: \*\*\* $p < 0.01$; \*\* $p < 0.05$; \* $p < 0.1$.*

**Appendix. Measurements**

Please indicate your experience and perceptions when you use the provided search tool for doing the above tasks based on a 7-points scale from strongly disagree to strongly agree.

*(Note: for the items with (-), we code them in a reverse way for analyses.)*

**Information quality**
1) I see a lot of not good or useless information in the responses. (-)
2) There is just too much information in the responses. (-)
3) What I am looking for during the search does not seem to be available. (-)

**Technology trust**
4) I fully rely on the answers from the provided tool in completing the tasks.
5) I feel I count on the provided tool when working on the tasks.
6) I think the answers from the provided tool are reliable.
7) I do not know if I can trust the answers from the provided tool. (-)

**Perceived ease of use**
8) I find completing the tasks using the provided tool easily.
9) I feel skillful in doing the tasks by the provided tool.
10) My interaction with the provided tool is clear and understandable.
11) I find the provided tool easy to use.
12) I find using the provided tool is cognitively demanding. (-)

**Perceived usefulness**
13) Using the provided tool enabled me to accomplish the tasks quickly.
14) Using the provided tool enhanced my effectiveness in completing the tasks.
15) I find the provided tool useful.

**Perceived enjoyment**
16) Using the provided tool is enjoyable.
17) I have fun with the provided tool.
18) I find using the provided tool interesting.

**Satisfaction**
19) I am satisfied with the use of the provided tool.
20) I am pleased to use the provided tool.
21) I like using the provided tool.

**Manipulation check questions:**

22) I used a tool that is similar to a traditional search engine. (-)
23) I used a tool with a conversation interface.